\documentclass[11pt,a4paper]{article}
\usepackage[hyperref]{emnlp2020}
\usepackage{times}
\usepackage{latexsym}
\usepackage{amsmath}

\usepackage{microtype}

\aclfinalcopy 

\usepackage{todonotes}
\usepackage{comment}

\definecolor{deepblue}{rgb}{0.16, 0.32, 0.75}
\definecolor{indiagreen}{rgb}{0.00, 0.44, 0.00}
\title{Event Presence Prediction Helps Trigger Detection Across Languages}

\author{Parul Awasthy \and
  Tahira Naseem \and Jian Ni \and Taesun Moon \and Radu Florian\\
    IBM Research AI\\
    Yorktown Heights, NY 10598\\
    {\{awasthyp, tnaseem, nij, tsmoon, raduf\}@us.ibm.com}\\
  }

\date{}
\begin{document}
\maketitle
\begin{abstract}

The task of event detection and classification is central to most information retrieval applications. We show that a Transformer based architecture can effectively model event extraction as a sequence labeling task. We propose a combination of sentence level and token level training objectives that significantly boosts the performance of a BERT based event extraction model. Our approach achieves a new state-of-the-art performance on ACE 2005 data for English and Chinese. We also test our model on ERE Spanish, achieving an average gain of 2 absolute $F_1$ points over prior best performing models. 

\end{abstract}

\begin{figure*}
\begin{center}
\includegraphics[scale=0.450]{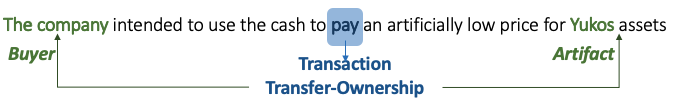}
\caption{Example of an Event.}\label{figure:example}
\end{center}
\end{figure*}

\section{Introduction and Prior Work}
Event Extraction (EE) is an important subfield in Information Extraction. It usually comprises two subtasks: (1) Event Detection (ED) which extracts from text nuggets of information that signals occurrence of an event; (2) Event Argument Extraction (EAE) which extracts  the participants of an event. The task is challenging as there are various ways to express the same event, and the same phrase can express different events. For example, \textit{pay} could mean event \textit{Transaction}, as in figure \ref{figure:example}, or could mean \textit{Justice}, as in ``...pay a fine of...". 

Events have been used by applications like Question Answering \cite{yang-et-al-qa}, Information Retrieval \cite{basile-et-al-2014}, Narrative Schema extraction \cite{10.5555/1690219.1690231}, Reading Comprehension \cite{cheng2018implicit}, and to build knowledge resources.

Traditionally, EE models were feature based, with careful feature engineering to generate rich set of features \cite{ahn-2006-stages,ji-grishman-2008-refining,li2013joint,araki-mitamura-2015-joint}. With the rise of deep learning based methods, recent works have used CNNs \cite{yubo2015event,lin-etal-2018-nugget}, RNNs \cite{nguyen2016joint}, LSTMs \cite{hong2018self,feng2018language}, Tree LSTMs \cite{yu-et-all-treelstm}, Graph CNNs \cite{nguyen2018graph} to extract events. DMBERT \cite{wang2019adversarial} was the first transformer based model. Since then \citet{wang2019adversarial,m'hamdi_etal_cross,Lin_OneIE} have also used transformers for EE. 

Work in other languages has been largely language dependent \cite{chen2009language}. Chinese EE models have relied on features or language characteristics to improve performance \cite{chen2012joint, lin-etal-2018-nugget}. \citet{feng2018language} and \citet{Lin_OneIE} have developed language independent models on English, Chinese and Spanish.

The best performing EE models rely on advanced architectures or language specific features.
Here, we propose a Transformer based model called sepBERT
which uses multiple objectives to improve performance.
The idea of learning additional objectives that exploit labels that are available with the data is used successfully by \citet{mikolov2013distributed, Devlin2018BERTPO, logeswaran2018efficient}, etc. 

We focus on Event Detection (ED) and
model the ED problem as a sequence labeling task. We introduce a simple and intuitive sentence level objective that greatly improves model performance. We show that for ED:
\begin{itemize}
    \item Adding an auxiliary objective function helps achieving the state-of-the-art performance.
    \item Highly contextualized Transformer models can be effectively used for ED without additional features or layers on top, performing better than more elaborate Transformer based models.
    \item Our model can be applied to different languages and can still achieve the state-of-the-art performance.
\end{itemize}


\section{Event Extraction Task}
\label{sec:task}
Event Extraction is defined for Automatic Content  Extraction (ACE) evaluation~\cite{walker2006ace} as follows:

\begin{itemize}
    \item Event Trigger is a phrase that  most  clearly  expresses the occurrence of an event. It is denoted by a type and subtype.
    \item Event Arguments are the mentions (entity, value or time) that serve as a participant or attribute with a specific role in the event. They are denoted by a role.
\end{itemize}

As shown in figure \ref{figure:example}, \textit{pay} is an event trigger of type \textbf{Transaction} and subtype \textbf{Transfer-Ownership} with two arguments \textit{The company} with role \textbf{Buyer} and \textit{Yukos} with role of \textbf{Artifact}. We focus on Event Trigger Extraction, also referred to as Event Detection (ED) in the literature. 


\section{Methodology}
\label{sec:method}

Most of the previous work on Trigger Extraction has focused on one word triggers \cite{nguyen2018graph,li2013joint,yubo2015event}, etc. Such methods classify each word of the sentence as a trigger, ignoring multi-word triggers. \citet{Lin_OneIE} predicts the start and end positions of the triggers to capture multi-word triggers. We treat the task as a sequence labelling task to implicitly capture multi-word triggers, as is done by \cite{m'hamdi_etal_cross}. 

\subsection{Trigger Extraction}
\begin{figure}[t]
\includegraphics[scale=0.55]{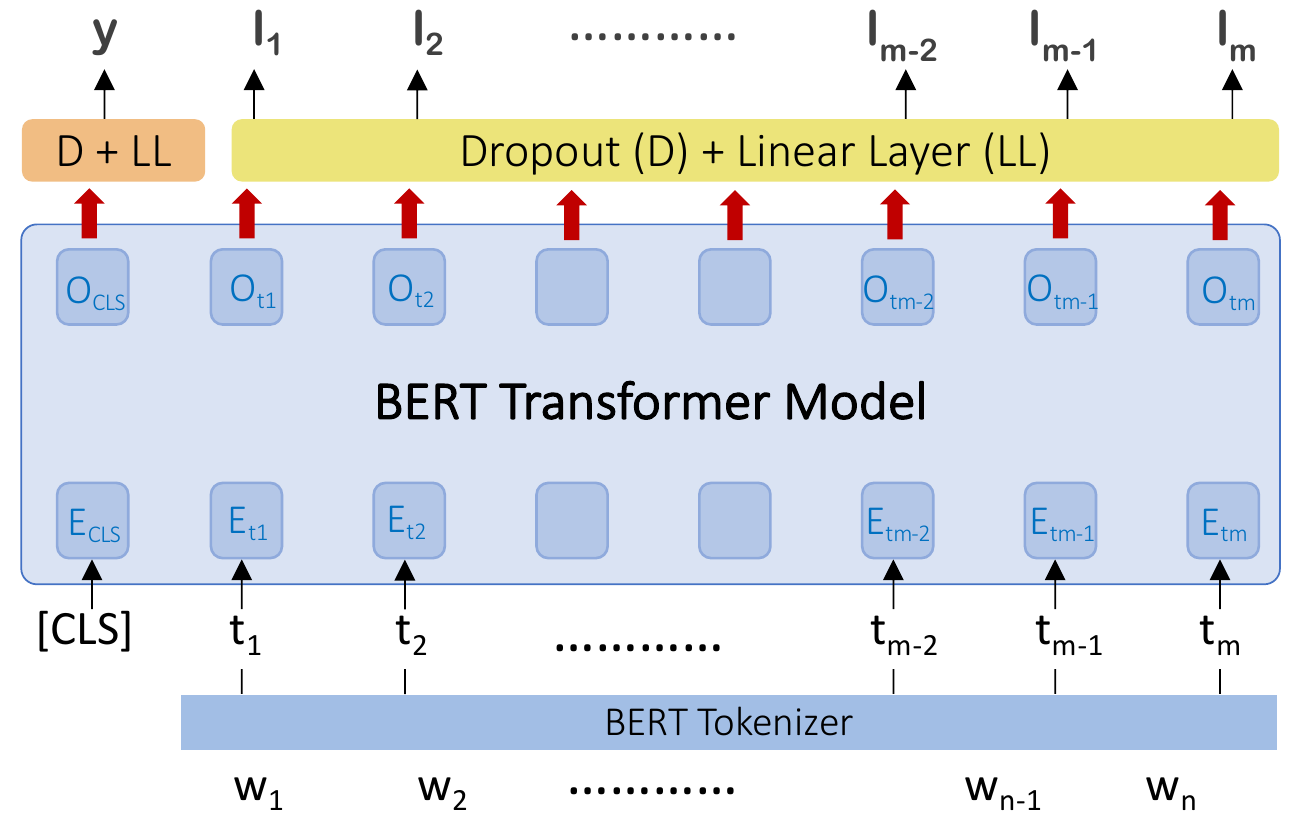}
\caption{Overall Architecture of sepBERT.}\label{figure:arch}
\end{figure}

Figure \ref{figure:arch} shows the architecture of our system, which builds on top of a pre-trained BERT model. We model trigger extraction as a token sequence labeling task. We combine the token-level classification loss $L_t$ with a novel sentence-level \emph{event presence} prediction loss $L_{sep}$: 
\vspace*{-1mm}\begin{equation*}
L = L_t + L_{sep}
\end{equation*}
For the token-level component of our model, we use the IOB2 encoding \cite{tjong-kim-sang-1999} where each token is labeled with its trigger label $l$ and an indicator of whether it starts or continues a trigger. We introduce a linear classification layer on top of the token-level BERT output vectors. The parameters of the layer are trained via cross entropy loss, a standard approach for BERT based sequence labeling models \cite{Devlin2018BERTPO}. This is equivalent to minimizing the negative log-likelihood of the true labels,

\begin{equation*}
L_{t} = -\sum_{i=1}^{n}{\log(P(l_{w_i}))}\nonumber
\end{equation*}
\noindent Although each token is classified independently, the BERT transformer layers help capture the context across all tokens via a multi-head self-attention mechanism. 

In addition, we introduce a simple yet effective sentence-level objective that serves as an auxiliary task. This is a binary classification objective, where truth $y = 1$ if the sentence contains an event and  $y=0$ otherwise. We model this by adding a classification layer over the BERT output of the [CLS] token -- a token added to the beginning of all BERT inputs,

\begin{equation*}
L_{sep} = -\log(P(y))\nonumber
\end{equation*}
 As many sentences do not have events, this loss acts as a discriminator. It offers the system a global representation of when events show up in sentences. We call this loss sentence event presence (SEP) loss and our model sepBERT. 

\section{Experiments}

\subsection{Dataset}
For English and Chinese we conduct our experiments on ACE~\cite{walker2006ace}, a popular event extraction dataset. ACE defines 8 event types and 33 subtypes. 

As ACE does not have an official test split, for English we follow the work of \cite{li2013joint} and use their pre-defined split of the documents to create a test set with the same 40 newswire documents, development set of 30 mixed documents and training set of the remaining 529. For Chinese, we follow the split proposed by \cite{feng2018language}, and randomly split the data into 60 development, 60 test, and the remaining 513 training documents.

For our Spanish language experiments we use the Entities, Relations and Events(ERE) corpus from LDC2015E107 catalog. ERE defines 9 event types, and 38 subtypes. We follow \cite{Lin_OneIE}'s split to split the 154 documents into 10 dev, 10 test and 134 train documents.


\subsection{Experimental Setup}
Details of our setup are mentioned below.
For contextualized word embedding, we use the HuggingFace PyTorch implementation of Transformers \cite{Wolf2019}. We train monolingual models on three languages, English, Spanish and Chinese, using the same method as described in \ref{sec:method}. We denote the language-specific models as sepBERT$_{lang}$. We use the out-of-the-box pre-trained transformer models, and fine-tune them with the event data, updating all layers.
Our English model uses \textit{bert-base-uncased} and the Chinese and Spanish model uses \textit{bert-base-multilingual-cased} with the standard BERT hyperparameters; we ran 20 epochs with 5 seeds each, dropout of 0.3, learning rate of $3\cdot 10^{-5}$ or $5\cdot 10^{-5}$, and training batch sizes of 30 for English, and 15 for Arabic and Spanish.   
We train our models on various seeds and learning rate combinations, and select the model that performs the best on the development set and report the test score for that model. Our sepBERT model takes about 2 minutes per epoch to train on a k80 GPU.

\subsubsection{Metrics}
We adopt the formal ACE evaluation criteria to evaluate our model: A trigger is correctly \emph{classified} if its offsets, event type, and subtype match those of a reference trigger. We use Precision, Recall and F$_1$-Measure to score the performance.

\subsection{Baselines}
\label{sec:baseline}
We create an ablation baseline, stdBERT, where we formulate the task as sequence labeling, but train without the SEP objective. This is a standard out-of-the-box sequence labeling model.

We compare our model with the following external \textbf{English} baselines:

\emph{Feature-based:} \citet{li2013joint}  is the best performing feature-based model. 
    
\emph{Neural net, pre-transformer:} JRNN \cite{nguyen2016joint} is a joint RNN based system. BiLSTM+GAN \cite{hong2018self} is an ED only NN model. GCN-ED \cite{nguyen2018graph} is a joint model that uses graph CNN to model dependency trees to capture event information.
    
\emph{Transformer-based:} DMBERT+Boot \cite{wang2019adversarial} uses \textit{bert-base-uncased} and models GANs to generate more training data for ED. They also have a vanilla baseline DMBERT, which does not use additional training data. BERT-CRF \cite{m'hamdi_etal_cross} is an ED only transformer based model that uses a CRF layer on top of \textit{bert-base} model. ONEIE \cite{Lin_OneIE} is a joint transformer based model, that extracts mentions, events and relations at the same time. They use \textit{bert-large-cased} for the English model. We report their score on what they call ACE-E+ dataset, as that matches our data.

\begin{table*}
\begin{centering}
\small
\begin{tabular}{|l|c|c|c|c|c|c|c|c|c|}
\hline 
 & \multicolumn{3}{c|}{English} & \multicolumn{3}{c|}{Chinese} & \multicolumn{3}{c|}{Spanish}\tabularnewline
\hline 
System & P & R & F  & P & R & F  & P & R & F \tabularnewline
\hline 
\small{}{Rich-C \cite{chen2012joint}} & -- & -- & -- & 58.9 & 68.1 & 63.2 & -- & -- & --\tabularnewline
\hline 
{\small{}{HNN \cite{feng2018language}} } & -- & -- & -- & 77.1 & 53.1 & 63.0 & -- & -- & --\tabularnewline
\hline 
{\small{}{NPN \cite{lin-etal-2018-nugget}}} & -- &  & -- & 60.9 & 69.3 & 64.8 & -- & -- & --\tabularnewline
\hline 
{\small{}{\cite{li2013joint}} } & 73.7 & 62.3 & 67.5 & -- & -- & -- & -- & -- & --\tabularnewline
\hline 
{\small{}{JRNN \cite{nguyen2016joint}$^J$} } & 66.0 & 73.0 & 69.3 & -- & -- & -- & -- & -- & --\tabularnewline
\hline 
{\small{}{BiLSTM+GAN \cite{hong2018self}$^J$} } & 71.3 & 74.7 & 73.0 & -- & -- & -- & -- & -- & --\tabularnewline
\hline 
{\small{}{GCN-ED\cite{nguyen2018graph}} } & 77.9 & 68.8 & 73.1 & -- & -- & -- & -- & -- & --\tabularnewline
\hline 
{\small{}{DMBERT \cite{wang2019adversarial}} } & 77.6 & 71.8 & 74.6 & -- & -- & -- & -- & -- & --\tabularnewline
\hline 
{\small{}{DMBERT+Boot \cite{wang2019adversarial}} } & 77.9 & 72.5 & 75.1 & -- & -- & -- & -- & -- & --\tabularnewline
\hline 
{\small{}{BERT-CRF \cite{m'hamdi_etal_cross}} } & -- & -- & 75.3 & -- & -- & 79.8{*} & -- & -- & --\tabularnewline
\hline 
{\small{}{ONEIE \cite{Lin_OneIE}$^J$} } & -- & -- & 72.8 & -- & -- & 65.6 & -- & -- & 56.8\tabularnewline
\hline 
\hline 
{\small{}{stdBERT$_{lang}$ } } & 68.3 & 79.8 & 73.6 & 63.7  & 66.6 & 65.1 & 59.0 & 62.3 & 60.6\tabularnewline
\hline 
{\small{}{sepBERT$_{lang}$} } & 74.2 & 78.5 & \textbf{76.3} & 67.8  & 65.6 & \textbf{66.7} & 59.55 & 64.7 & \textbf{62.0}\tabularnewline
\hline 
\end{tabular}
\par\end{centering}
\caption{Trigger Classification Performance on English, Chinese, and Spanish. stdBERT$_{lang}$ and sepBERT$_{lang}$ are trained on monolingual $lang$ data;  ${^J}$ is used to differentiate Joint EE models.}

\label{table:all-trigc} 
\end{table*}

For \textbf{Chinese}, we compare our model with the following external baselines:

\emph{Language dependent:} Rich-C \cite{chen2012joint} is a feature-based model that adds language specific features to improve performance. NPN \cite{lin-etal-2018-nugget} is a Chinese only Neural Net model that learns a hybrid representation for each character from both characters and words.

\emph{Language independent:} HNN \cite{feng2018language} is a BiLSTM based Neural Net model. ONEIE \cite{Lin_OneIE} is a transformer based model that uses \textit{bert-base-multilingual-cased}.

*BERT-CRF* \cite{m'hamdi_etal_cross} although we have listed BERT-CRF Chinese language scores, their numbers are not comparable to ours as they ``limit the maximum sequence  length of sentences to 128, padding or cutting otherwise." 
This leads to reduction in the ground truth making their total event count 2521, when there are 3333 total events in the Chinese data set. 
\footnote{It is to be noted that their English event count does not suffer from this problem, possibly due to the shorter length of ACE05 English sentences.}

Our \textbf{Spanish} baseline is ONEIE.

\section{Results and Discussion}

Table \ref{table:all-trigc} shows our results. The sepBERT model consistently achieves state-of-the-art results across languages. Some baselines systems, annotated with $^J$ in Table \ref{table:all-trigc} are joint event extraction model, optimizing a global objective over event triggers and arguments. 
Our model outperforms them and the Event Detection only models like DMBERT, BERT-CRF, BiLSTM+GAN with an average gain more than 2F points.

On English, our ablation baseline stdBERT$_{en}$, performs better than the feature-based model and the JRNN model by a good margin. It performs at par with the GCN-ED and BiLSTM+GAN, but does worse than the tranformer based baselines. Our full model sepBERT$_{en}$ performs 1F point better than BERT-CRF, and 1.8F points better than DMBERT, both of which are transformer based methods. It performs 1.2F points better than DMBERT+Boot, system that augments the training data, and more than 3F points better than the more elaborate neural net models GCN-ED and BiLSTM-GAN. It is 3.5F points better than ONEIE, which is a larger model built using \textit{bert-large}. 

Similarly, on Chinese our ablation baseline stdBERT$_{zh}$ does better than all other models, except transformer based ONEIE. sepBERT$_{zh}$ is again the best model, performing 1.1F points better than ONEIE, also based on \textit{bert-base-multilingual}, and 1.9+F points better than language specific models like NPN and Rich-C. It is also 3.7F points better than the other language independent baseline HNN. *As mentioned in \ref{sec:baseline} BERT-CRF is not comparable with our model. 

On Spanish sepBERT$_{es}$ is 5.2F points better than ONEIE. The stdBERT$_{es}$ also scores 3.8F higher than ONEIE.

Results show that the stdBERT baseline has a lower precision than sepBERT. By comparing their output we see that stdBERT$_{en}$ produces 30\% more false positives than sepBERT$_{en}$. This suggests that the added event presence objective helps make sepBERT$_{en}$ more precise. To further investigate, we measure the sentence level event prediction accuracy of both systems, i.e. we measure ratio of number of times system is correct in predicting presence of an event with total number of times a system predicts any event in a sentence. We find for all languages, sepBERT is more accurate than stdBERT, by as much as 3 points, suggesting the SEP loss works as desired.

We also find that both sepBERT$_{en}$ and stdBERT$_{en}$ mis-classify around 5\% of the events, i.e., they label a trigger of type x as y, where both x,y are valid trigger types and not "O". This suggests that the added objective does not make the model more accurate on the identified events.

\section{Conclusion}
In this paper, we proposed a method for event trigger extraction (sepBERT), which formulates the task as sequence labeling to capture multi-word triggers. With the help of an added auxiliary objective, sepBERT achieves the state-of-the-art performance. We show that our model effectively uses the pre-trained Transformer without additional features or layers on top, performing better than more elaborate Transformer based models. Moreover, our model can be applied to different languages and can still achieve the state-of-the-art performance in those languages, attaining an average gain of 2 absolute $F_1$ points over past best performing models.

\bibliographystyle{acl_natbib}
\bibliography{emnlp2020.bib}

\begin{thebibliography}{26}
\expandafter\ifx\csname natexlab\endcsname\relax\def\natexlab#1{#1}\fi

\bibitem[{Ahn(2006)}]{ahn-2006-stages}
David Ahn. 2006.
\newblock \href {https://www.aclweb.org/anthology/W06-0901} {The stages of
  event extraction}.
\newblock In \emph{Proceedings of the Workshop on Annotating and Reasoning
  about Time and Events}, pages 1--8, Sydney, Australia. Association for
  Computational Linguistics.

\bibitem[{Araki and Mitamura(2015)}]{araki-mitamura-2015-joint}
Jun Araki and Teruko Mitamura. 2015.
\newblock \href {https://doi.org/10.18653/v1/D15-1247} {Joint event trigger
  identification and event coreference resolution with structured perceptron}.
\newblock In \emph{Proceedings of the 2015 Conference on Empirical Methods in
  Natural Language Processing}, pages 2074--2080, Lisbon, Portugal. Association
  for Computational Linguistics.

\bibitem[{Basile et~al.(2014)Basile, Caputo, Semeraro, and
  Siciliani}]{basile-et-al-2014}
Pierpaolo Basile, Annalina Caputo, Giovanni Semeraro, and Lucia Siciliani.
  2014.
\newblock Extending an information retrieval system through time event
  extraction.
\newblock \emph{CEUR Workshop Proceedings}, 1314:36--47.

\bibitem[{Chambers and Jurafsky(2009)}]{10.5555/1690219.1690231}
Nathanael Chambers and Dan Jurafsky. 2009.
\newblock Unsupervised learning of narrative schemas and their participants.
\newblock In \emph{Proceedings of the Joint Conference of the 47th Annual
  Meeting of the ACL and the 4th International Joint Conference on Natural
  Language Processing of the AFNLP: Volume 2 - Volume 2}, ACL ’09, page
  602–610, USA. Association for Computational Linguistics.

\bibitem[{Chen and Ng(2012)}]{chen2012joint}
Chen Chen and Vincent Ng. 2012.
\newblock Joint modeling for chinese event extraction with rich linguistic
  features.
\newblock In \emph{Proceedings of COLING 2012}, pages 529--544.

\bibitem[{Chen et~al.(2015)Chen, Xu, Liu, Zeng, and Zhao}]{yubo2015event}
Yubo Chen, Liheng Xu, Kang Liu, Daojian Zeng, and Jun Zhao. 2015.
\newblock \href {https://doi.org/10.3115/v1/P15-1017} {Event extraction via
  dynamic multi-pooling convolutional neural networks}.
\newblock In \emph{Proceedings of the 53rd Annual Meeting of the Association
  for Computational Linguistics and the 7th International Joint Conference on
  Natural Language Processing (Volume 1: Long Papers)}, pages 167--176,
  Beijing, China. Association for Computational Linguistics.

\bibitem[{Chen and Ji(2009)}]{chen2009language}
Zheng Chen and Heng Ji. 2009.
\newblock Language specific issue and feature exploration in chinese event
  extraction.
\newblock In \emph{Proceedings of Human Language Technologies: The 2009 Annual
  Conference of the North American Chapter of the Association for Computational
  Linguistics, Companion Volume: Short Papers}, pages 209--212.

\bibitem[{Cheng and Erk(2018)}]{cheng2018implicit}
Pengxiang Cheng and Katrin Erk. 2018.
\newblock Implicit argument prediction with event knowledge.
\newblock \emph{arXiv preprint arXiv:1802.07226}.

\bibitem[{Devlin et~al.(2019)Devlin, Chang, Lee, and
  Toutanova}]{Devlin2018BERTPO}
Jacob Devlin, Ming-Wei Chang, Kenton Lee, and Kristina Toutanova. 2019.
\newblock {BERT}: Pre-training of deep bidirectional transformers for language
  understanding.
\newblock In \emph{NAACL-HLT}.

\bibitem[{Feng et~al.(2018)Feng, Qin, and Liu}]{feng2018language}
Xiaocheng Feng, Bing Qin, and Ting Liu. 2018.
\newblock A language-independent neural network for event detection.
\newblock \emph{Science China Information Sciences}, 61(9):092106.

\bibitem[{Hong et~al.(2018)Hong, Zhou, Zhang, Zhou, and Zhu}]{hong2018self}
Yu~Hong, Wenxuan Zhou, Jingli Zhang, Guodong Zhou, and Qiaoming Zhu. 2018.
\newblock Self-regulation: Employing a generative adversarial network to
  improve event detection.
\newblock In \emph{Proceedings of the 56th Annual Meeting of the Association
  for Computational Linguistics (Volume 1: Long Papers)}, pages 515--526.

\bibitem[{Ji and Grishman(2008)}]{ji-grishman-2008-refining}
Heng Ji and Ralph Grishman. 2008.
\newblock \href {https://www.aclweb.org/anthology/P08-1030} {Refining event
  extraction through cross-document inference}.
\newblock In \emph{Proceedings of ACL-08: HLT}, pages 254--262, Columbus, Ohio.
  Association for Computational Linguistics.

\bibitem[{Li et~al.(2013)Li, Ji, and Huang}]{li2013joint}
Qi~Li, Heng Ji, and Liang Huang. 2013.
\newblock Joint event extraction via structured prediction with global
  features.
\newblock In \emph{Proceedings of the 51st Annual Meeting of the Association
  for Computational Linguistics (Volume 1: Long Papers)}, pages 73--82.

\bibitem[{Lin et~al.(2018)Lin, Lu, Han, and Sun}]{lin-etal-2018-nugget}
Hongyu Lin, Yaojie Lu, Xianpei Han, and Le~Sun. 2018.
\newblock \href {https://doi.org/10.18653/v1/P18-1145} {Nugget proposal
  networks for {C}hinese event detection}.
\newblock In \emph{Proceedings of the 56th Annual Meeting of the Association
  for Computational Linguistics (Volume 1: Long Papers)}, pages 1565--1574,
  Melbourne, Australia. Association for Computational Linguistics.

\bibitem[{Lin et~al.(2020)Lin, Ji, Huang, and Wu}]{Lin_OneIE}
Ying Lin, Heng Ji, F~Huang, and L~Wu. 2020.
\newblock A joint neural model for information extraction with global features.
\newblock In \emph{Proc. The 58th Annual Meeting of the Association for
  Computational Linguistics (ACL2020)}.

\bibitem[{Logeswaran and Lee(2018)}]{logeswaran2018efficient}
Lajanugen Logeswaran and Honglak Lee. 2018.
\newblock An efficient framework for learning sentence representations.
\newblock \emph{arXiv preprint arXiv:1803.02893}.

\bibitem[{M{'}hamdi et~al.(2019)M{'}hamdi, Freedman, and
  May}]{m'hamdi_etal_cross}
Meryem M{'}hamdi, Marjorie Freedman, and Jonathan May. 2019.
\newblock \href {https://doi.org/10.18653/v1/K19-1061} {Contextualized
  cross-lingual event trigger extraction with minimal resources}.
\newblock In \emph{Proceedings of the 23rd Conference on Computational Natural
  Language Learning (CoNLL)}, pages 656--665, Hong Kong, China. Association for
  Computational Linguistics.

\bibitem[{Mikolov et~al.(2013)Mikolov, Sutskever, Chen, Corrado, and
  Dean}]{mikolov2013distributed}
Tomas Mikolov, Ilya Sutskever, Kai Chen, Greg~S Corrado, and Jeff Dean. 2013.
\newblock Distributed representations of words and phrases and their
  compositionality.
\newblock In \emph{Advances in neural information processing systems}, pages
  3111--3119.

\bibitem[{Nguyen et~al.(2016)Nguyen, Cho, and Grishman}]{nguyen2016joint}
Thien~Huu Nguyen, Kyunghyun Cho, and Ralph Grishman. 2016.
\newblock Joint event extraction via recurrent neural networks.
\newblock In \emph{Proceedings of the 2016 Conference of the North American
  Chapter of the Association for Computational Linguistics: Human Language
  Technologies}, pages 300--309.

\bibitem[{Nguyen and Grishman(2018)}]{nguyen2018graph}
Thien~Huu Nguyen and Ralph Grishman. 2018.
\newblock Graph convolutional networks with argument-aware pooling for event
  detection.
\newblock In \emph{Thirty-second AAAI conference on artificial intelligence}.

\bibitem[{Sang and Veenstra(1999)}]{tjong-kim-sang-1999}
Erik F. Tjong~Kim Sang and Jorn Veenstra. 1999.
\newblock \href {https://doi.org/10.3115/977035.977059} {Representing text
  chunks}.
\newblock In \emph{Proceedings of the Ninth Conference on European Chapter of
  the Association for Computational Linguistics}, EACL ’99, page 173–179,
  USA. Association for Computational Linguistics.

\bibitem[{Walker et~al.(2006)Walker, Strassel, Medero, and
  Maeda}]{walker2006ace}
Christopher Walker, Stephanie Strassel, Julie Medero, and Kazuaki Maeda. 2006.
\newblock \href {https://catalog.ldc.upenn.edu/LDC2006T06} {Ace 2005
  multilingual training corpus}.
\newblock \emph{Linguistic Data Consortium, Philadelphia}, 57.

\bibitem[{Wang et~al.(2019)Wang, Han, Liu, Sun, and Li}]{wang2019adversarial}
Xiaozhi Wang, Xu~Han, Zhiyuan Liu, Maosong Sun, and Peng Li. 2019.
\newblock Adversarial training for weakly supervised event detection.
\newblock In \emph{Proceedings of the 2019 Conference of the North American
  Chapter of the Association for Computational Linguistics: Human Language
  Technologies, Volume 1 (Long and Short Papers)}, pages 998--1008.

\bibitem[{Wolf et~al.(2019)Wolf, Debut, Sanh, Chaumond, Delangue, Moi, Cistac,
  Rault, Louf, Funtowicz, and Brew}]{Wolf2019}
Thomas Wolf, Lysandre Debut, Victor Sanh, Julien Chaumond, Clement Delangue,
  Anthony Moi, Pierric Cistac, Tim Rault, Rémi Louf, Morgan Funtowicz, and
  Jamie Brew. 2019.
\newblock \href {https://arxiv.org/abs/1910.03771} {Huggingface's transformers:
  State-of-the-art natural language processing}.
\newblock \emph{ArXiv}, abs/1910.03771.

\bibitem[{Yang et~al.(2003)Yang, Chua, Wang, and Koh}]{yang-et-al-qa}
Hui Yang, Tat-Seng Chua, Shuguang Wang, and Chun-Keat Koh. 2003.
\newblock \href {https://doi.org/10.1145/860435.860444} {Structured use of
  external knowledge for event-based open domain question answering}.
\newblock In \emph{SIGIR '03}, pages 33--40.

\bibitem[{Yu et~al.(2020)Yu, Yi, Huang, Yi, and Yuan}]{yu-et-all-treelstm}
Wentao Yu, Mianzhu Yi, Xiaohui Huang, Xiaoyu Yi, and Qingjun Yuan. 2020.
\newblock \href {https://doi.org/10.1109/ACCESS.2020.2965964} {Make it
  directly: Event extraction based on tree- lstm and bi-gru}.
\newblock \emph{IEEE Access}, PP:1--1.

\end{thebibliography}

\end{document}